# Graph Based Link Prediction between Human Phenotypes and Genes

Rushabh Patel[1], Yanhui Guo[2]

[1]Dept.of Computer & Information Science, Temple University, Philadelphia, USA
[2]University of Illinois Springfield, Springfield, IL USA

**Abstract**

**Background**
The learning of genotype-phenotype associations and history of human disease by doing detailed and precise analysis of phenotypic abnormalities can be defined as deep phenotyping. To understand and detect this interaction between phenotype and genotype is a fundamental step when translating precision medicine to clinical practice. The recent advances in the field of machine learning is efficient to predict these interactions between abnormal human phenotypes and genes.

**Methods**
In this study, we developed a framework to predict links between human phenotype ontology (HPO) and genes. The annotation data from the heterogeneous knowledge resources i.e., orphanet, is used to parse human phenotype-gene associations. To generate the embeddings for the nodes (HPO & genes), an algorithm called node2vec was used. It performs node sampling on this graph based on random walks, then learns features over these sampled nodes to generate embeddings. These embeddings were used to perform the downstream task to predict the presence of the link between these nodes using 5 different supervised machine learning algorithms.

**Results:**
The downstream link prediction task shows that the Gradient Boosting Decision Tree based model (LightGBM) achieved an optimal AUROC 0.904 and AUCPR 0.784. In addition, LightGBM achieved an optimal weighted F1 score of 0.87. Compared to the other 4 methods LightGBM is able to find more accurate interaction/link between human phenotype & gene pairs.

## 1. Introduction

Today, many humans have diseases with abnormalities in the genome and due to their nonuniformity, many diseases are undiagnosed. The analysis of human phenotype plays a key role in medical research & clinical practice [1]. The HPO is able to play an important role by translating precision medicine into clinical practice in deep phenotyping. Deep phenotyping can be defined as an in-depth analysis of phenotypic abnormalities with precision in which phenotype component for an individual is observed and described [2]. The Human Phenotype Ontology (HPO) is commonly used as a resource that systematically defines and logically organizes human phenotypes [3]. The HPO is collected majorly from medical literature and different knowledge resources, including a database of chromosomal imbalance and human phenotypes using DECIPHER [4], OMIM [5], and the Orphanet [6]. The majority of the studies in this field identify relevant phenotypes mainly based on topological & ancestorial relationships of any two nodes in the directed acyclic graph. These studies do not take feature learning through embeddings of different nodes to detect associations which is difficult to infer through graph structure. The traditional classification methods were applied to predict the interactions and associations between HPO-gene terms. These methods introduced inconsistencies because target values are predicted without considering inherited relationships within the ontology. For instance, if we are trying to predict the link between human phenotype-gene, a normal classifier will associate the HPO term "Squamous Cell Carcinoma" with a gene, but it might not associate "Abnormality of the Skin", hence leading to an inaccurate prediction. To appropriately handle the hierarchical relationships between HPO terms that accurately characterize HPO, we use node embeddings. The node embeddings provide a solution to map graph nodes to distributed representations & allowing it to translate the relationship from graph to embedding space. The commonly adopted method to build node embeddings is Node2Vec [7]. It performs a flexible neighborhood sampling strategy using biased random walk and passing this sampling data to the word2vec model as input [8]. In this study, we are designing a framework to predict the interaction between human phenotype-gene. The dataset used for this study was taken from the heterogeneous knowledge resources specifically the Orphanet. We have converted this knowledge resource data to an undirected graph and using this graph to create node embeddings to learn the features. These node features were used to build downstream machine learning models to predict interactions. We conducted quantitative analysis on the output of these models using different metrics.

## 2. Related Work:

In the clinical service domain, several studies have used the Human Phenotype Ontology graph structure to understand the association & interaction between different phenotype ontology, genes & proteins. HPO2GO [9] study shows ways to predict associations between human phenotype terms in conjunction with genes [10]. OntoFUNC [11] integrated pharmacogenomics databases to identify associations between chemical pathways using analysis on chemical ontology. HPOAnnotator [12] infer large-scale protein-human phenotype association using PPI (Protein-Protein Interaction) information & the HPO Graph. They use low-rank approximation to solve the sparsity problem in finding associations. However, none of these methods uses node embeddings & advanced machine learning models to extract appropriate node features for predicting human phenotype-gene associations.

## 3. Methodology:

### 3.1 Data Preparation:

Link prediction aims to find missing links or identifying future link interactions between nodes based on the currently observed partial network. Predicting links between nodes has been the most important research topic in the field of graphs and networks. To use machine learning for predicting the interaction between the unconnected node-pairs of the graph, we need to represent the graph in the form of a structured dataset having a set of features. In figure 1, we have a graph with 7 nodes and unconnected node pairs AF, BD, BE, BG, EG. Let's suppose we analyze the graph data at time *t* and found new connections that have been formed in the graph at time *t+n* (red links) in Fig. 2. We extract unconnected node pairs A-F, B-D, B-E, B-G, E-G, and then we look at the graph at time *t+n* and label these three new links (red lines in Fig 2) in the graph for the node pairs A-F, B-D & B-E as 1 and the node-pairs BG & EG will be labeled as 0 because still at time *t+n* no links were found for these nodes.

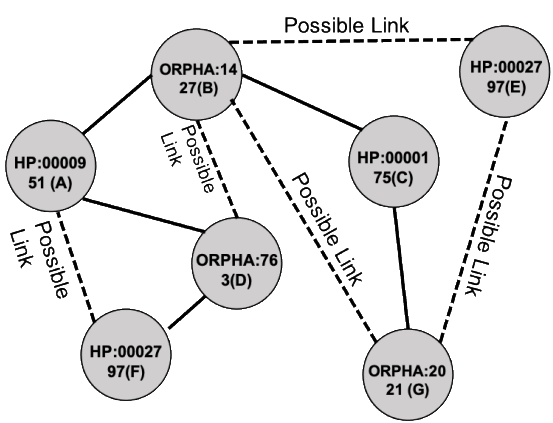 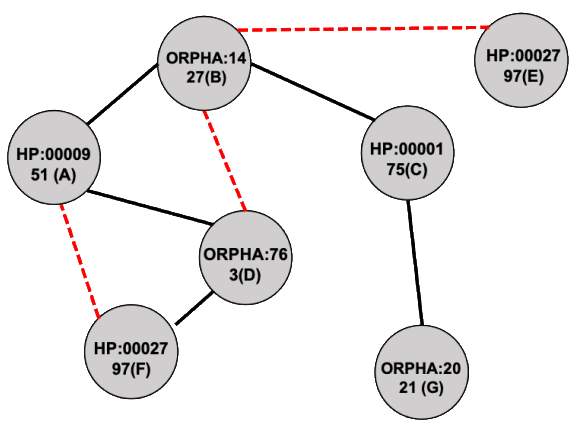

**Figure 1**
*Graph at time t*

**Figure 2**
*Graph at time t+n*

In this example, we had access to the graph at time *t+n,* and that is the reason we were able to get the labels for the target variable. However, in real-world networks or graphs, we would have access to a single large graph. We need to first understand that connections between nodes in the graph are built gradually over time and hence keeping this in mind, randomly hiding some of the edges in the given graph and then creating labels would solve our problem. But, while removing links or edges, we should always avoid removing edges that may lead to an isolated node i.e., a node without any edge or an isolated network.

A major part of the undirected graph obtained from the Orphanet HPO annotation dataset belongs to negative population or unconnected node-pairs. To find these node pairs we create an adjacency matrix as shown in Figure 3. This adjacency matrix is a square matrix where both rows and columns are defined by nodes of the graph. The value in the matrix denotes edges or links. The value of one means edge exists and zero means no edge found between that node-pair. In Fig. 3, nodes HP:0000951 and ORPHA:763 have value 0 at a cross junction in the adjacency matrix and there is also no edge or link between them in the graph.

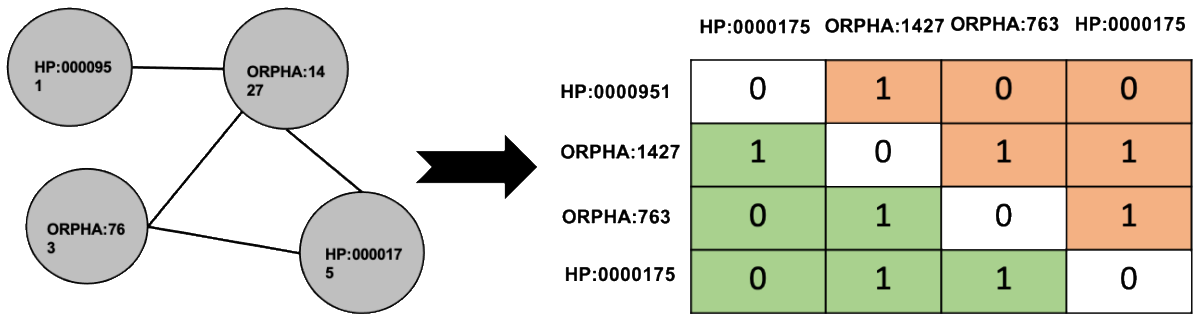

**Figure 3**
*Finding unconnected node pairs using Adjacency Matrix*

In this square matrix, we will iterate either above or below the diagonal because both provide similar information, and it will be redundant to traverse through the whole matrix. The traversal will help us to find the positions of the negative samples. Similarly, we randomly drop few edges to get the positive samples but doing so may drop loosely connected fragments & nodes. To handle this, we first check for the splitting of the graph while dropping a node-pair or checking on the counts for total nodes. If these conditions are met, then we can safely drop that node pair. By following these steps, we found a total of 1665146 unconnected node pairs and out of which only 125304 positive samples i.e., around 7.5% (highly imbalanced dataset). The heterogeneous knowledge resource, Orphanet contains 133733 associations between human phenotype & genes. The statistics for the undirected graph generated from this data can be found in Figure 4.

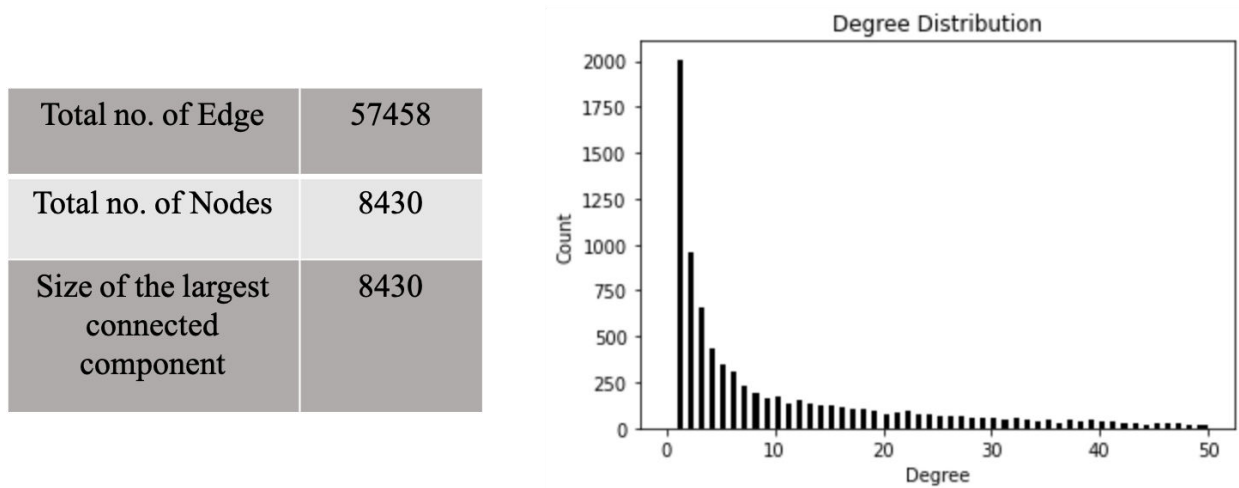

| | |
|---|---|
| Total no. of Edge | 57458 |
| Total no. of Nodes | 8430 |
| Size of the largest connected component | 8430 |

**Figure 4**
*Graph Statistics & Degree Distribution*

### 3.2 Feature Extraction:

To extract features from graph G, we use the node2vec algorithm. It creates vector space embeddings to represent node features. Node2Vec starts with the weighted random walks from every node of the graph and interprets these random walks as terms which can be embedded by using a skip gram model into Euclidean Space **[13].** The aim of this approach is to maximize the node *n*'s probability in context *K* within the contextual window of length l:

$$L(n) = \sum_{k \in K} \sum_{i=1}^{|k|} \sum_{-l \leq j \leq l, i \neq j} \log p(k_i \mid k_j) \qquad (1)$$

Here $k_i$ denotes the $i^{th}$ term or node sequence generated by using random walk. The output from the skip gram method which is being defined by the SoftMax function is *p(k_j | k_i )*

$$p(k_i \mid k_j) = \frac{exp - (\langle n_i, n_j \rangle)}{\sum_{-l \leq j \leq l} exp - (\langle n_i, n_j \rangle)} \qquad (2)$$

where $n_j$, $n_i$ are vector representations of terms $k_j$, $k_i$ in the hidden layer of the skip gram model. The node2vec library was used to train a model with 30 nodes in each walk and 200 walks per node with 128 embedding dimensions. We will apply this *node2vec* model to every node pair in the dataset.

## 3.3 Downstream Prediction using ML algorithms:

We use extracted node features from the node2vec algorithm as an input to machine learning models. To evaluate the performance of these algorithms, we split the complete dataset into training (80%) & testing (20%). We used the following methods for this supervised link prediction task.

### 3.3.1 Logistic Regression:

Logistic Regression is a widely used method in classifying binary outcomes borrowed from the field of statistics. It uses the sigmoid activation function to restrict the outcome between 0 & 1. The coefficients are estimated from our training set using a maximum-likelihood learning algorithm. It is a common algorithm that makes assumptions about the distribution of data. For this study, we have used with L-BFGS solver[14].

### 3.3.2 Random Forest:

Random Forest is a widely used bagging method for supervised machine learning tasks. The decision trees are ensembled together to generate forest trained using bagging methods. The bagging here refers to build multiple decision trees together and later merging them to get a more accurate prediction. This algorithm can be used for both classifications as well as a regression task. In our case, we will be training the Random Forest model for binary classification to predict the link between two nodes. The maximum depth used to train this model was 12 [15].

### 3.3.3 Neural Network:

Neural networks have revolutionized the field of machine learning with recent advancements. This algorithm tries to mimic the human brain to identify patterns in the data. Neural networks can be used for a wide variety of tasks including clustering similar data, classifying different objects, and many more. The neural network is initiated with random weights & thresholds. The training data is being fed as vectors to the input layer and letting it pass through the succeeding hidden layers by complexly adjusting these weights & thresholds until we yield similar outputs as true labels. For this study, we are using Neural Network with two hidden layers each with the relu activation and using sigmoid activation at the output layer to get the output values between 0 and 1. We used Adam optimizer with a learning rate value of 1e-3 which is an extension to SGD (Stochastic Gradient Descent) which helps us to update the network weight based on training data. [16].

### 3.3.4 XGBoost:

XGBoost (e**X**treme **G**radient **Boost**ing) is a go-to algorithm in the machine learning community to solve most of the classification & regression problems. It uses boosting technique by training models in an isolation from each other by trying to correct mistakes from the previous models. These models are added up sequentially until we reach the point where no further mistakes can be made. Gradient Boosting is a method which trains on the errors or residuals of previous models. This method is computationally efficient & fast as compared to other boosting methods. For this study, we have used the learning rate of 0.1, maximum depth of 12, and scale weight of 0.99 to handle the imbalanced nature of the data [17].

### 3.3.5 LightGBM

LightGBM is one of the gradient boosting methods which is very fast & computationally efficient. This framework is used for many downstream prediction tasks like classification, regression & ranking. It uses the same concepts as XGBoost except for one key difference, it splits trees on the leaf and not on the level. This helps to reduce more loss when growing on the same leaf. We use maximum depth parameter value 10 to avoid building a more complex model. The scale pos weight 99 was used to deal with the class imbalance problem for this study.

### 4. Results and Discussion

### 4.1 Evaluation Metrics

Due to high imbalance, the model evaluation metrics we choose are important. In this study, we try to understand how each class is performing rather than focusing on overall metrics. Below are the metrics which we have used to validate the performance of these machine learning methods.

### 4.1.1 AUROC:

AUROC stands for Area Under the Receiver Operating Characteristics. In a binary classification task, a very common summary statistic to calculate the goodness of a predictor is defined by AUROC. In AUROC, the Receiver Operating Characteristics is the probability curve, and Area

Under Curve is the degree of separability. It provides information about the class distinguishability in the model. This ROC curve is plotted with True Positive Rate (Sensitivity) against the False Positive Rate (1 - Specificity).

$$Sensitivity\ (TPR) = \frac{True\ Positives}{True\ Positive + False\ Negatives} \quad (3)$$

$$False\ Positive\ Rate\ (FPR) = \frac{False\ Positives}{False\ Positive + True\ Negatives} \quad (4)$$

### 4.1.2 AUCPR:

AUCPR stands for the Area Under the Precision-Recall Curve. This is the most widely used metric in the machine learning community for validating highly imbalanced data. This is used for evaluating binary classification model performance by plotting Precision against Recall. These metrics help us to see the performance of positive samples more closely as compared to AUROC.

$$Precision = \frac{True\ Positives}{True\ Positives + False\ Positives} \quad (5)$$

$$Recall\ (Sensitivity) = \frac{True\ Positives}{True\ Positives + False\ Negatives} \quad (6)$$

### 4.1.3 Micro Average Precision/Recall/F1 Score:

The micro average metric helps us to weigh each class instance equally.
**PrecisionMicroAvg** can be defined as the sum of all True Positives for all classes by all positive predictions.

$$PrecisionMicroAvg = \frac{(TP1+TP2+\cdots+TPn)}{(TP1+TP2+\cdots+TPn+FP1+FP2+\cdots+FPn)} \quad (7)$$

**RecallMicroAvg** can be defined as the sum of all True Positives for all classes by all actual positives i.e., *True Positives* not the predicted positives.

$$RecallMicroAvg = \frac{(TP1+TP2+\cdots+TPn)}{(TP1+TP2+\cdots+TPn+FN1+FN2+\cdots+FNn)} \quad (8)$$

**F1MicroAvg** can be defined by calculating the harmonic mean for **PrecisionMicroAvg** & **RecallMicroAvg**.

$$F1MicroAvg = 2 * \frac{PrecisionMicroAvg * RecallMicroAvg}{PrecisionMicroAvg + RecallMicroAvg} \quad (9)$$

### 4.1.4 Macro Average Precision/Recall/F1 Score:

The macro average is used when we want to treat all classes equally when evaluating performance [18] concerning the most frequent class in the dataset.

**PrecisionMacroAvg** can be defined as the arithmetic mean of precision from all the classes. It can be mathematically defined as

$$PrecisionMacroAvg = \frac{(Precision_1 + Precision_2 + \cdots + Precision_n)}{n} \quad (10)$$

**RecallMacroAvg** can be defined as the arithmetic mean of recall from all the classes. It can be mathematically defined as

$$RecallMacroAvg = \frac{(Recall_1 + Recall_2 + \cdots + Recall_n)}{n} \quad (11)$$

**F1MacroAvg** can be defined as the mean of the F1 score by class wise.

$$F1MacroAvg = \frac{1}{N}\sum_{i=0}^{N} F1_{class\ i} \quad (12)$$

where $i$ is the class index & $N$ is the total number of classes.

### 4.1.5 Weighted Average Precision/Recall/F1 Score:

**PrecisionWeightedAvg** can be calculated by multiplying class weights based on true labels with its corresponding precision score.

$$PrecisionWeightedAvg = \frac{(Precision_1 * W_1 + Precision_2 * W_1 + \cdots + Precision_n * W_n)}{n} \quad (13)$$

where $W$ is class weight based on true labels.

Similarly, **RecallWeightedAvg** can be calculated by multiplying class weights based on true labels with its corresponding recall score.

$$RecallWeightedAvg = \frac{(Recall_1 * W_1 + Recall_2 * W_1 + \cdots + Recall_n * W_n)}{n} \quad (14)$$

where $W$ is class weight based on true labels.

**F1WeightedAvg** can be calculated by multiplying class weights based on true labels with its corresponding f1 score. Mathematically, it can be represented as below.

$$F1WeightedAvg = \frac{1}{N}\sum_{i=0}^{N} F1_{class\ i} * W_{class\ i} \qquad (15)$$

where **W** is class weight based on true labels.

## 4.2 Comparing Performances of different models.

To evaluate the predictive performance of these five models, Logistic Regression, Random Forest, Neural Network, XGBoost & LightGBM on human phenotype-gene dataset we use all the many different metrics including AUROC, AUCPR, Micro, Macro & Weighted Average precision, recall & F1 score, *etc*. To appropriately evaluate the imbalance nature of the dataset we calculate important metrics for each class instance in our case we just have two classes 0 and 1.

**Table 1** *Class-wise Evaluation*

| Model | Class | Precision | Recall | F1 score |
|---|---|---|---|---|
| Logistic Regression | 0 | 0.96 | 0.71 | 0.81 |
|  | 1 | 0.14 | 0.60 | 0.23 |
| Random Forest | 0 | 0.94 | 1.00 | 0.97 |
|  | 1 | 1.00 | 0.18 | 0.30 |
| Neural Network | 0 | 0.98 | 0.87 | 0.92 |
|  | 1 | 0.34 | 0.78 | 0.47 |
| XGBoost | 0 | 0.95 | 1.00 | 0.97 |
|  | 1 | 0.99 | 0.34 | 0.50 |
| LightGBM | 0 | 0.98 | 0.84 | 0.91 |
|  | 1 | 0.30 | 0.82 | 0.44 |

As you can see from Table 1, it is much easier to identify which model is doing a great job in identifying each class. Based on these metrics we can see that XGBoost & Random Forest performs well in identifying positive samples who are positives i.e., When these predict a link between nodes, they are correct 99% & 100% of the time respectively. Contrastingly, LightGBM beats all other methods in identifying correct actual positive – in other words, it correctly predicts 82% of all the links between these nodes. From Table 2, we can confer that LightGBM is better than all other models in terms of AUROC & AUCPR.

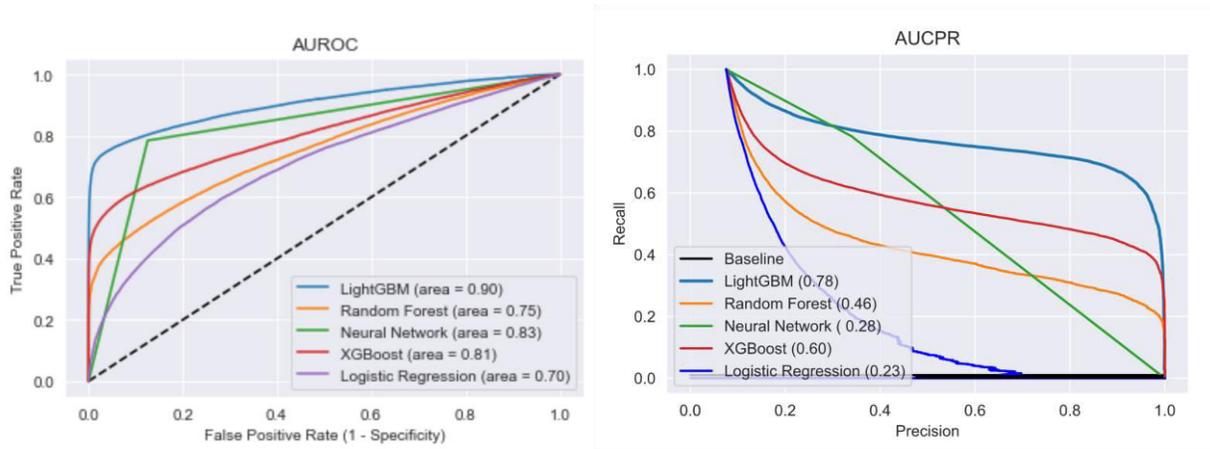

**Figure 5** *AUCROC & AUCPR*

Table 2 *Micro, Macro & Weighted Metrics*

| Model | Metric | Micro | Macro | Weighted | AUROC | AUCPR |
|---|---|---|---|---|---|---|
| Logistic Regression | Precision | 0.70 | 0.55 | 0.89 | 0.70 | 0.23 |
| | Recall | 0.70 | 0.66 | 0.70 | | |
| | F1 score | 0.70 | 0.52 | 0.77 | | |
| Random Forest | Precision | 0.94 | 0.97 | 0.94 | 0.75 | 0.46 |
| | Recall | 0.94 | 0.59 | 0.94 | | |
| | F1 score | 0.94 | 0.63 | 0.92 | | |
| Neural Network | Precision | 0.87 | 0.66 | 0.93 | 0.83 | 0.28 |
| | Recall | 0.87 | 0.83 | 0.87 | | |
| | F1 score | 0.87 | 0.70 | 0.89 | | |
| XGBoost | Precision | 0.95 | 0.97 | 0.95 | 0.81 | 0.60 |
| | Recall | 0.95 | 0.67 | 0.95 | | |
| | F1 score | 0.95 | 0.74 | 0.94 | | |
| LightGBM | Precision | 0.84 | 0.64 | 0.93 | 0.90 | 0.78 |
| | Recall | 0.84 | 0.83 | 0.84 | | |
| | F1 score | 0.84 | 0.67 | 0.87 | | |

## 5. Conclusion

In this study, we presented an approach to predict links between human phenotype & genes using heterogeneous knowledge resources i.e., orphanet. The most important part of this study is to represent data into a graph and then finding a way to represent this graph into an appropriate feature set which will allow us to use it for down streaming tasks like a prediction. In essence, we provided a way to get the embedding vectors by using an algorithm called node2vec and then using these embeddings to build five different machine learning models. We evaluated and compared the performances using different quantitative metrics including AUROC, AUCPR, Micro, Macro & Weighted Precision, Recall, and F1 score. Some of these metrics were calculated for each class instance to better understand the situation for imbalanced class, in our case positive samples. Based on these metrics we found very interesting results. If we want to just focus on positive samples meaning the measure of the link that we correctly identify having associations of all the actual associations in the graph (we refer to it as Precision), then we may either use XGBoost or Random Forest algorithm. On the other hand, if we just want to focus on accurately identifying positives from True Positives i.e., actual links in the graph (Recall) then use LightGBM.